\documentclass[doublecolumn,letterpaper,10pt,conference]{ieeeconf}

\let\labelindent\relax

\usepackage{breakcites}
\usepackage{amsmath,amssymb,amsfonts}
\usepackage{graphicx}
\usepackage{textcomp}
\usepackage{styles/style}
\usepackage{indentfirst}
\usepackage{rotating}
\usepackage{algorithm,algorithmic}
\usepackage{pgf}
\usepackage{tikz}
\usepackage{tikz-3dplot}
\usepackage{pifont}
\usepackage{aligned-overset}
\usepackage{longtable}
\usepackage{dsfont}
\usepackage{booktabs}
\usepackage{nicefrac}
\usepackage{amsmath}
\usepackage[usenames,dvipsnames]{pstricks}
\usetikzlibrary{arrows,shapes,automata,decorations.text,decorations.pathmorphing,backgrounds,positioning,fit}
\usepackage{enumitem}
\usepackage{color}
\usepackage{xcolor}
\usepackage{tabularx}
\usepackage{pgfplots}
\usepackage{pst-grad} 
\usepackage{pst-plot} 
\usetikzlibrary{positioning}
\usepackage{rotating}
\usepackage{cite}
\usepackage[colorlinks=true, allcolors=blue]{hyperref}
\usepackage{dblfloatfix}
\usepackage{url,amsmath,setspace,amssymb}
\usepackage{tikz}
\usepackage{graphicx}
\usepackage{caption}
\usepackage{subcaption}
\usepackage{float}
\usepackage{soul}
\usepackage{epsfig}
\usepackage{longtable}
\usepackage{relsize}
\allowdisplaybreaks
\IEEEoverridecommandlockouts
\usepackage{nccmath}

\usepackage{xargs}                      
\usepackage[colorinlistoftodos]{todonotes}

\usetikzlibrary{external}
\definecolor{royalblue}{RGB}{65, 105, 225}
\definecolor{seagreen}{RGB}{46, 139, 87}
\definecolor{firebrick}{RGB}{178,34,34}
\definecolor{darkviolet}{RGB}{138, 43, 226}
\definecolor{carrotorange}{RGB}{237, 145, 33}
\pgfplotsset{every tick label/.append style={font=\small}}

\def\BibTeX{{\rm B\kern-.05em{\sc i\kern-.025em b}\kern-.08em
    T\kern-.1667em\lower.7ex\hbox{E}\kern-.125emX}}
\begin{document}
\title{\LARGE \bf On First-Order Meta-Reinforcement Learning with Moreau Envelopes}
\author{Mohammad Taha Toghani$^\dagger$,
Sebastian Perez-Salazar$^\star$,
C\'{e}sar A. Uribe$^\dagger$
\thanks{
$^\dagger$Department of Electrical and Computer Engineering, $^\star$Computational Applied Mathematics and Operations Research, Rice University, Houston, TX, USA. Email addresses: \{\href{mailto:mttoghani@rice.edu}{mttoghani},
\href{mailto:sperez@rice.edu}{sperez}, \href{mailto:cauribe@rice.edu}{cauribe}\}@rice.edu.}}

\allowdisplaybreaks

\maketitle

\begin{abstract}
Meta-Reinforcement Learning (\mtext{MRL}) is a promising framework for training agents that can quickly adapt to new environments and tasks. In this work, we study the \mtext{MRL} problem under the policy gradient formulation, where we propose a novel algorithm that uses Moreau envelope surrogate regularizers to jointly learn a meta-policy that is adjustable to the environment of each individual task. Our algorithm, called Moreau Envelope Meta-Reinforcement Learning (\mtext{MEMRL}), learns a meta-policy that can adapt to a distribution of tasks by efficiently updating the policy parameters using a combination of gradient-based optimization and Moreau Envelope regularization. Moreau Envelopes provide a smooth approximation of the policy optimization problem, which enables us to apply standard optimization techniques and converge to an appropriate stationary point. We provide a detailed analysis of the \mtext{MEMRL} algorithm, where we show a sublinear convergence rate to a first-order stationary point for non-convex policy gradient optimization. We finally show the effectiveness of \mtext{MEMRL} on a multi-task $2$D-navigation problem.
\end{abstract}


\section{Introduction}
\label{sec:introduction}

The Reinforcement Learning (\mtext{RL}) problem~\cite{szepesvari2010algorithms,sutton2018reinforcement,agarwal2019reinforcement,liu2021temporal} studies the interaction of some agent with an environment to maximize a reward function. In this problem setup, the agent observes the environment's state, selects an action according to a policy, receives a reward and a new state, and updates its policy based on experience \cite{szepesvari2010algorithms}. Policy Gradient Reinforcement Learning (\mtext{PGRL})~\cite{sutton1999policy} is a subclass of \mtext{RL} methods that directly optimize the policy by following the gradient of the expected reward with respect to the policy parameters~\cite{van2019use}. A neural network or another parametric function usually represents the policy, mapping states to action probabilities~\cite{clifton2020q}. \mtext{PGRL} methods are advantageous because they can handle high-dimensional and continuous action spaces and integrate prior knowledge or structure into the policy~\cite{schulman2015high}.

Meta-Reinforcement Learning (\mtext{MRL}) focuses on enabling agents to learn how to learn or adapt rapidly to new tasks and environments~\cite{finn2017model,rajeswaran2019meta,finn2019online}. In \mtext{MRL}, the agent is trained to perform (i.e., a higher reward in expectation) not just on a single task but on a range of tasks, each defined by distinct reward functions, initial state distributions, and transition dynamics~\cite{nagabandi2018learning}. Thus, the agent learns a meta-policy that can quickly adapt to new tasks by modifying its policy or value function based on a few experience samples from the new task~\cite{finn2017model,dorfman2021offline}. The meta-policy aims to optimize the expected cumulative reward across tasks rather than maximizing the reward for a single task. The \mtext{MRL} framework enables agents to generalize better to new tasks and contexts, making it a crucial area of research.

The primary motivation behind \mtext{MRL} is the need for agents to adapt rapidly to frequently changing environments or tasks, allowing them to become more efficient~\cite{dasgupta2019causal,schoettler2020meta,wang2020offline}. For instance, a robot that operates in various environments~\cite{schoettler2020meta} or a recommender system that adapts to different user preferences~\cite{wang2020offline}. \mtext{MRL} equips agents with the ability to generalize their skills and knowledge across tasks, allowing them to learn new tasks more efficiently by leveraging past experiences. By enabling agents to learn how to learn, \mtext{MRL} opens new possibilities for intelligent systems to adapt and succeed in complex and dynamic environments.

Some works discuss the fast adaptation in \mtext{MRL} via different techniques~\cite{ren2022efficient,melo2022transformers,zintgraf2022fast}. Ren et al.~\cite{ren2022efficient} develop an algorithm that can adapt to new tasks with preference-based feedback from a human oracle. The algorithm uses information theory techniques to design query sequences that maximize the information gained from human interactions while tolerating the inherent error of a non-expert human oracle. Melo~\cite{melo2022transformers} presents a method that uses the transformer architecture to mimic the memory reinstatement mechanism. The agent associates the recent past of working memories to build episodic memory recursively through the transformer layers. Zintgraf~\cite{zintgraf2022fast} proposes an \mtext{MRL} framework that can adapt to new tasks by learning a latent task representation and a task-conditioned policy. The framework uses variational inference techniques to infer the task representation from trajectories and optimize the policy with respect to task distribution.

The main challenge in \mtext{MRL} is finding a good representation of the tasks and meta-policy that enables efficient adaptation to new tasks while preserving knowledge learned from past tasks. Beck et al.~\cite{beck2023survey}  provide a comprehensive overview of the \mtext{MRL} problem setting, its variations, algorithms, and applications, along with the open challenges for this problem. Furthermore, Yu et al.~\cite{yu2020meta} present a benchmark suite for \mtext{MRL} that includes robotic manipulation tasks with varying difficulty and diversity.

\mtext{MRL} algorithms use different techniques such as gradient-based methods~\cite{finn2017model}, model-based~\cite{clavera2018model} or model-free \mtext{RL}~\cite{li2021provably}, and shared hierarchy (or representation)~\cite{frans2017meta,zhang2021learning} to tackle these challenges. Frans et al.~\cite{frans2017meta} present \mtext{MetaSH}, an \mtext{MRL} algorithm that learns a shared hierarchy of policies that can generalize across tasks. This method uses a high-level policy that selects sub-policies based on the task context and a low-level policy that executes actions based on the sub-policy. Finn et al.~\cite{finn2017model} introduce \mtext{MAML}, another \mtext{MRL} algorithm that learns a model-agnostic initialization that can be fine-tuned with a few gradient steps to new tasks. \mtext{MAML} can be applied to any model trained with gradient descent/ascent and loss functions.

The main challenge faced by Model-Agnostic Meta-Reinforcement Learning (\mtext{MAMRL}) and its variants~\cite{finn2017model,fallah2021convergence,rajeswaran2019meta} is the scalability in training. The \mtext{MAMRL} formulation requires computing and storing multiple gradients and Hessians for each task and updating the meta-policy with respect to these parameters. This may incur high computational and memory costs, especially for large-scale or continuous problems~\cite{shin2021large}. Moreover, the stability of training, which depends on the number of gradient steps, poses another challenge for \mtext{MAMRL}-based methods. Such issues affect the convergence and generalization of \mtext{MAMRL} and require careful tuning/trade-off. Some possible solutions are to use Hessian-free methods~\cite{fallah2020convergence} or adaptive learning rates~\cite{jiang2019improving}.

This work studies the \mtext{MRL} problem through gradient-based techniques. We formulate the \mtext{MRL} problem via surrogate cost functions, i.e., Moreau Envelope proximal operators~\cite{moreau1965proximite,parikh2013proximal} and derive the first-order information for this framework building upon the policy gradient problem setup. Our main contribution is \textit{lowering memory requirements} and \textit{reducing arithmetic complexity} by removing the need for second-order information, i.e., a reduction from $\mcO(d^2)$ to $\mcO(d)$. We list our contributions as follows:
\begin{itemize}[leftmargin=2em]
    \item We propose a novel framework for \mtext{MRL} via Moreau Envelopes and propose a novel first-order algorithm called \mtext{MEMRL} to maximize the proposed personalized value function.
    \item We present a detailed convergence analysis of the proposed algorithm under customary assumptions and show a sublinear convergence result for our proposed method.
    \item We finally present the performance of \mtext{MEMRL} on multi-task $2$D-navigation on a discrete grid with a set of finite actions.
\end{itemize}

The remainder of this paper is arranged as follows. In Section~\ref{sec:setup}, we introduce the underlying problem setup, compare with prior works, and present our novel algorithm \mtext{MEMRL} for Meta-Reinforcement Learning with Moreau Envelopes. In Section~\ref{sec:convergence}, we present the convergence result of our proposed algorithm along with the underlying assumptions. We provide a numerical experiment demonstrating the performance of our method in Section~\ref{sec:experiments}. Finally, we conclude the remarks and highlight future works in Section~\ref{sec:conclusion}.


\section{Problem Setup \& Algorithm}\label{sec:setup}
In this section, we first describe the Meta-Reinforcement Learning (\mtext{MRL}) problem setup and discuss the Policy Gradient Reinforcement Learning (\mtext{PGRL}) method via function approximation. Then, we explain the setup for \mtext{MRL} through Moreau Envelope auxiliary cost and a brief comparison with some relevant works. Finally, we present our method \mtext{MEMRL} for the underlying problem.

\subsection{Policy Gradient Meta-Reinforcement Learning} 

We consider a set of (potentially infinite) Markov Decision Processes (MDPs) \mbox{$\{\mcM_i\}_{i\in\mcI}$} that represent different tasks drawn from a distribution $p$ over a finite time horizon\footnote{A common assumption in \mtext{RL} is that the agent operates in an infinite-horizon setting, where the goal is to maximize the expected discounted or average reward over an infinite number of steps. However, in many practical scenarios, the agent may face a finite-horizon setting, where the goal is to maximize the expected discounted reward over a finite number of steps~\cite{vp2021finite}.} \mbox{$\{0, 1, \dots, H\}$}. For each task \mbox{$i\in\mcI$}, we denote the states and actions by $\mcS_i$ and $\mcA_i$, respectively. In this setup, the initial state distribution is given by \mbox{$\mu_i:\mcS_i\to\Delta(\mcS_i)$}, where \mbox{$\Delta(S_i)$} is the set of probability distributions over $\mcS_i$. Moreover, the transition kernel is denoted by $\mcP_i$, where \mbox{$\mcP_i(s'_i|s_i, a_i)$} is the probability of transitioning from state \mbox{$s_i\in\mcS_i$} to \mbox{$s'_i\in\mcS_i$} by taking action \mbox{$a_i\in\mcA_i$} for which a reward \mbox{$r_i(s_i, a_i)$} is received according to its corresponding reward function \mbox{$r_i: \mcS_i{\times} \mcA_i \to [0, R]$}. Therefore, the value of a trajectory \mbox{$\tau_i=(s_i^0, a_i^0, \dots,a_i^{H{-}1},s_i^H)$} can be defined as
\begin{align}\label{eq:full-reward}
    \mcR_i(\tau_i) \coloneqq \sum_{h{=}0}^{H{-}1} \gamma^h r_i(s_i^h,a_i^h),
\end{align}
where \mbox{$\gamma\in(0,1)$} is the discount factor for reward accumulation over time. Eventually, each task \mbox{$i\in\mcI$} can be modeled as an MDP defined by the tuple \mbox{$(\mcS_i, \mcA_i, \mcP_i, r_i, \mu_i, \gamma)$}. In this setting, a random policy \mbox{$\pi_i:\mcS_i \to \Delta(\mcA_i)$} determines the probability of each action $a_i$ given a state $s_i$ as \mbox{$\pi_i(a_i|s_i)$}. In \textit{Policy Gradient Reinforcement Learning (\mtext{PGRL})}, we parameterize the policy by a $d$-dimensional parameter $w\in\bbR^d$ (like a large neural network), i.e., $\pi_i(\cdot|\cdot;w)$. Therefore, the probability of trajectory $\tau_i$ is given by
\begin{align}
    q_i(\tau_i;w)\coloneqq \mu_i(s_i^0)\prod_{h{=}0}^{H{-}1} \pi_i(a_i^h|s_i^h;w) \prod_{h{=}0}^{H{-}1} \mcP_i(s_i^{h{+}1}|s_i^h,a_i^h).
\end{align}
Accordingly, the average reward value for each task \mbox{$i\in\mcI$} is
\begin{align}\label{eq:cost-j-i}
    J_i(w) \coloneqq \bbE_{\tau_i \sim q_i(\cdot;w)}\left[\mcR_i(\tau_i)\right],
\end{align}
which is a function of parameter $w$. In multi-task \mtext{RL} problems, we seek to find a joint parameter that maximizes the expected reward on all tasks $\mcI$:
\begin{align}\label{eq:rl}
    J(w) \coloneqq \bbE_{i\sim p}\left[J_i(w)\right].
\end{align}
The goal of policy gradient is to find a (sub)optimal parameter that maximizes the expected cumulative reward in~\eqref{eq:rl} obtained by $\pi_i(\cdot|\cdot;w)$, for all \mbox{$i\in\mcI$}. The key idea behind \mtext{PGRL} is to use the gradient of value function with respect to the policy parameters $w$ to update the policy in the direction that increases the expected cumulative reward.

In multi-task settings with heterogeneous environments, we seek to find a global policy \mbox{$w\in\bbR^d$} that performs well by adapting quickly to each task, i.e., obtaining a personal policy \mbox{$\theta_i\in\bbR^d$} through fine-tuning. We formulate the joint multi-task setup via \textit{Moreau Envelope Meta-Reinforcement Learning} cost (\mtext{MEMRL})
\begin{subequations}\label{eq:memrl}
\begin{align}
    \max_{w\in\bbR^d} V(w) &\coloneqq \bbE_{i\sim p}\left[V_i(w)\right]\label{eq:memrl-global}\\
    \text{with}\quad V_i(w) &\coloneqq \max_{\theta_i\in\bbR^d}\left[J_i(\theta_i) - \frac{\lambda}{2}\lVert \theta_i-w\rVert^2\right]\label{eq:memrl-local},
\end{align}
\end{subequations}
where parameter $\lambda{\geq}0$ forms a trade-off on the similarity of policies for different tasks. The formulation in~\eqref{eq:memrl} is a bilevel optimization problem. A solution $w^\star\in\bbR^d$ to Problem~\eqref{eq:memrl-global} is considered a meta-model that yields a task-personalized parameter $\theta_i^\star$ by maximizing Problem~\eqref{eq:memrl-local}.
In the next subsection, we discuss the cost function in~\eqref{eq:memrl} and present a novel policy gradient-based algorithm with access to a first-order oracle to maximize this cost.

\noindent\textbf{Related Works:} The \mtext{MRL} problem has been studied~\cite{rajeswaran2019meta,finn2019online} for multi-task setups with parameter adjustment mainly under Model-Agnostic Meta-Learning (\mtext{MAML}) framework~\cite{finn2017model}, where the goal is to maximize the following cost function:
\begin{subequations}\label{eq:maml}
\begin{align}
    \max_{w\in\bbR^d} V'(w) &\coloneqq \bbE_{i\sim p}\left[V'_i(w)\right],\label{eq:maml-global}\\
    \text{with}\quad V'_i(w) &\coloneqq J_i(w + \alpha \nabla J_i(w)).\label{eq:maml-local}
\end{align}
\end{subequations}
This formulation suggests to find an initial policy that performs well after modification with one step of gradient ascent. Finn et al.~\cite{finn2019online} studied the online meta-learning setting under the \mtext{MAML} setup and Rajeswaran et al.~\cite{rajeswaran2019meta} established one of the first theoretical results for this framework. Some other recent works have examined the complexity analysis of \mtext{MAML} in different contexts such as supervised meta-learning~\cite{fallah2020convergence,ji2020multi}. Assuming the inner loop loss function is sufficiently smooth and strongly convex, \mtext{iMAML} converges to a first-order stationary point in the deterministic case. Moreover, works such as~\cite{fallah2021convergence,ji2020multi,toghani2022parspush} study the impact of multi-step fine-tuning, an extended version of~\eqref{eq:maml}, under appropriate assumptions. Specifically,~\cite{fallah2021convergence} establishes the first analysis for multi-step \mtext{MRL} with stochastic gradients via a novel algorithm called \mtext{SGMRL}.

Different variations of \mtext{MAML} setup require access to the second-order information in the underlying update rule for the policy gradient maximization. Moreover, in the \mtext{RL} setups, \mtext{MAML} with one gradient often fails to perform well in practice, hence the choice of multi-step \mtext{MAML} with multiple updates remains a crucial for \mtext{MRL} tasks. In the contrary, Moreau Envelope (\mtext{ME})~\cite{dinh2020personalized,toghani2022persafl} controls the trade-off between the similarity and divergence of personal parameters $\theta_i$ via the regularization parameter $\lambda$. Furthermore, the inner optimization problem can be maximized without the need for second-order information.


\subsection{Meta-Reinforcement Learning with Moreau Envelopes}\label{subsec:alg}
We start by presenting the first-order information of~\eqref{eq:cost-j-i}. 
According to~\cite{peters2008reinforcement,shen2019hessian,sutton2018reinforcement,fallah2021convergence}, the gradient of $J_i(\cdot)$ can be derived via the logarithm derivative trick as follows:
\begin{align}\label{eq:full-grad-j-i}
    \nabla J_i(w) \coloneqq \bbE_{\tau_i \sim q_i(\cdot;w)}\left[g_i(\tau_i;w)\right],
\end{align}
where the stochastic policy gradient $g_i(\cdot;w)$ is given by
\begin{subequations}\label{eq:stoch-grad-j-i-def}
\begin{align}
    g_i(\tau_i;w) &\coloneqq  \sum_{h=0}^{H{-}1} \nabla_{w} \log \pi_i(a_i^h|s_i^h;w) \,\mcR_i^h(\tau_i),\label{eq:stoch-grad-j-i}\\
    \text{where}\quad \mcR_i^h(\tau_i) &\coloneqq \sum_{l{=}h}^{H{-}1} \gamma^l\, r_i(s_i^l,a_i^l).\label{eq:partial-reward}
\end{align}
\end{subequations}
We drop parameter $w$ from the gradient notation for simplicity of presentation in~\eqref{eq:stoch-grad-j-i}. In this setup, $g_i(\cdot;w)$ is the score function that measures the sensitivity of the log-probability of the trajectory $\tau_i$ to the policy parameters $w$. Moreover,~\eqref{eq:full-reward} and~\eqref{eq:partial-reward} imply \mbox{$\mcR_i^0(\tau_i) = \mcR_i(\tau_i)$}.

To deal with the computational intractability of the full gradient in~\eqref{eq:full-grad-j-i}, we approximate this term by a stochastic policy gradient over a batch $\mcD_{i}$ of trajectories sampled from distribution $q_i(\cdot;w)$, i.e.,
\begin{align}\label{eq:batch-grad-j-i}
    \nabla \tilde{J}_i(\mcD_{i}; w) \coloneqq \frac{1}{|\mcD_{i}|} \sum_{\tau_i\in\mcD_{i}} g_i(\tau_i;w),
\end{align}
where $\nabla J_i(w) {=} \bbE\left[\nabla \tilde{J}_i(\mcD_{i}; w)\right]$. 

Next, we present the first-order information of the surrogate function $V_i(\cdot)$ in~\eqref{eq:memrl-local}. To compute the gradient, let
\begin{align}\label{eq:optimal-surrogate}
    \hat{\theta}_i(w) \coloneqq \argmax_{\theta_i\in\bbR^d}\left\{J_i(\theta_i) - \frac{\lambda}{2}\norm{\theta_i-w}^2\right\},
\end{align}
then, by taking derivative of $V_i(\cdot)$ with respect to $w$, we have
\begin{align}\label{eq:chain-rule}
    \nabla V_i(w) = \frac{\partial \hat{\theta}_i(w)}{\partial w}&\nabla J_i(\hat{\theta}_i(w))\nonumber\\
    &- \lambda\left[\frac{\partial \hat{\theta}_i(w)}{\partial w}-I\right]\left(\hat{\theta}_i(w)-w\right),
\end{align}
and due to first-order optimality,
\begin{align}
    \nabla J_i(\hat{\theta}_i(w)) - \lambda \left(\hat{\theta}_i(w)-w\right)&=0 \overset{\eqref{eq:chain-rule}}{\Rightarrow}\label{eq:grad-v-i-proof}\\
    \nabla V_i(w) &= \lambda \left(\hat{\theta}_i(w)-w\right).\label{eq:grad-v-i}
\end{align}
Therefore, one needs to obtain $\hat{\theta}_i(w)$ for computing $\nabla V_i(w)$. Note that given a set of parameters $w$, finding $\hat{\theta}_i(w)$ for a general policy function $\pi_i(\cdot|\cdot;w)$ is computationally intractable, since (i) deterministic gradient $\nabla J_i(\cdot)$ cannot be computed and (ii) an exact solution to~\eqref{eq:optimal-surrogate} requires certain assumptions on the function class, e.g., quadratic functions~\cite{charles2021convergence}. Hence, we instead propose to maximize the following stochastic approximation with respect to parameter~$\theta_i$
\begin{align}
\tilde{F}_i\left(\mcD_i;\theta_i,w\right) \coloneqq \tilde{J}_i\left(\mcD_i;\theta_i\right) - \frac{\lambda}{2}\left\lVert\theta_i - w\right\rVert^2,\label{eq:stoch-f-i}
\end{align}
where $\mcD_i$ is a batch of trajectories sampled from distribution $q_i(\cdot;\theta_i)$. Thus, it is sufficient to find an inexact solution $\tilde{\theta}_i(w)$ that satisfies
\begin{align}\label{eq:stoch-approx-error}
\norm{\nabla_{\theta_i}\tilde{F}_i\left(\mcD_i;\tilde{\theta}_i(w),w\right)}\leq \nu,
\end{align}
for some approximation precision $\nu>0$, where
\begin{align}
\nabla_{\theta_i}\tilde{F}_i\left(\mcD_i;\theta_i,w\right) = \nabla \tilde{J}_i\left(\mcD_i;\theta_i\right) - \lambda(\theta_i - w).\label{eq:stoch-grad-f-i}
\end{align}
Then, we can approximate the exact gradient $\nabla V_i(w)$ in~\eqref{eq:grad-v-i} with
\begin{align}
\nabla\tilde{V}_i(w)\coloneqq\lambda(\tilde{\theta}_i(w)-w),\label{eq:stoch-grad-v-i}
\end{align}
where $\tilde{\theta}_i(w)$ satisfies~\eqref{eq:stoch-approx-error}. Note that a small parameter $\nu$ provides a better approximation, thus less error in the solution of the algorithm.

\begin{algorithm}[t]
    \caption{\mtext{MEMRL}: First-Order \underline{M}oreau \underline{E}nvelope \underline{M}eta-\underline{R}einforcement \underline{L}earning}
    \begin{algorithmic}[1]
    \STATE{\textbf{input:} 
    regularization parameter $\lambda$, inexact approximation precision $\nu$, meta stepsize $\alpha$, task batch size $B$, trajectory batch size $D$.
    }
    \STATE{\textbf{initialize:} $w^0\in\bbR^d, t\leftarrow 0$}
    \REPEAT
        \STATE{sample a batch of tasks $\mcB^t\subseteq \mcI$ with size $B$}
        \FOR{all tasks $i\in\mcB^t$}
        \STATE{find $\tilde{\theta}_i(w^t)$ such that for a batch of trajectories $\mcD_i^t$ (of size $D$) sampled from $q_i(\cdot;\tilde{\theta}_i(w^t))$ to maximize $\tilde{F}_i\left(\cdot;\cdot,w^t\right)$ up to accuracy level $\nu$ with
        \begin{align*}
        \norm{\nabla\tilde{F}_i\left(\mcD^t_i;\tilde{\theta}_i(w^t),w^t\right)}\leq \nu
        \end{align*}}\label{step:approx}
        
        \vspace{-1em}
        \ENDFOR
        \STATE{$w^{t{+}1} \leftarrow (1{-}\alpha\lambda)w^t + \frac{\alpha\lambda}{|\mcB^t|} \sum\limits_{i\in\mcB^t}\tilde{\theta}_i(w^t)$}\label{step:aggregation}
        \vspace{-0.5em}
        \STATE{$t\leftarrow t + 1$}
    \UNTIL{not converged}
    \STATE{\textbf{output:} }
    \end{algorithmic}
    \label{alg:MEMRL}
\end{algorithm}

We are now ready to propose \mtext{MEMRL} for solving the problem in~\eqref{eq:memrl}. Algorithm~\ref{alg:MEMRL} shows the pseudo-code for our method. Starting from a random initial set of parameter $w^0$, we perform an iterative method. At each round $t\geq 0$, we sample a batch of tasks $\mcB^t$ with size $B$ and for each task $i\in\mcB^t$, we maximize $\nabla_{\theta_i}\tilde{F}_i\left(\cdot;\cdot,w^t\right)$ up to precision $\nu$ (Step~\ref{step:approx} of Algorithm~\ref{alg:MEMRL}). Then, we use the approximate individual sub-optimal solutions $\tilde{\theta}_i(w^t)$ to approximate the gradient of $\nabla V_i(w^t)$ according to \eqref{eq:stoch-grad-v-i} and use this to aggregate and apply one step of gradient ascent in Step~\ref{step:aggregation} of Algorithm~\ref{alg:MEMRL}.

The inexact optimization solver for the inner problem in Step~\ref{step:approx} of Algorithm~\ref{alg:MEMRL} can be any first-order policy gradient method. For example, let us describe an algorithm for the inexact optimizer:
\begin{enumerate}
    \item $\theta^{t,0}_{i}\leftarrow w^t, k\leftarrow 0$,
    \item sample a batch of trajectories $\mcD^{t,0}_i$ with size $D$ with respect to $q_i(\cdot;\theta^{t,0}_{i})$,
    \vspace{0.2em}
    \item While not $\norm{\nabla \tilde{F}_i\left(\mcD_{i}^{t,k};\theta^{t,k}_{i},w^t\right)}\leq\nu$:
    \begin{enumerate}[leftmargin=2em]
    \item sample a batch of trajectories $\mcD_i^{t,k}$ with size $D$ with respect to $q_i(\cdot;\theta^{t,k}_{i})$,
    \vspace{0.2em}
    \item $\theta^{t,k{+}1}_{i} \leftarrow \theta^{t,k}_{i} + \beta \Big[\nabla \tilde{J}_i(\mcD^{t,k}_i;\theta^{t,k}_{i}) - \lambda (\theta^{t,k}_{i}{-}w^t)\Big]$,
    \item $k \leftarrow k + 1$,
    \end{enumerate}
    \item $\tilde{\theta}_i(w^t) \leftarrow \theta^{t,k}_{i}$.
\end{enumerate}

We will discuss the convergence of Algorithm~\ref{alg:MEMRL} in the next section. 



\section{Convergence Result}\label{sec:convergence}
We start this section by stating the underlying assumption for our analysis. Further, we present two auxiliary lemmas stating the properties of $J_i$ and $V_i$ functions. Finally, we show the convergence result of \mtext{MEMRL} as the main result of this work along with its proof.

\begin{assumption}[Log-Probability Properties]\label{assump:main}
The logarithm of the policy functions $\pi_i$ are twice differentiable, for all $i\in\mcI$. Moreover,
there exist constants $G$ and $L$, such that for any task $i\in\mcI$ and state $s_i\in\mcS_i$, action $a_i\in\mcA_i$, and arbitrary parameter $w\in\bbR^d$,
\begin{align}
        \norm{\nabla \log \pi_i(a_i|s_i;w)}\leq G,\label{eq:bounded-grad}\\
        \norm{\nabla^2 \log \pi_i(a_i|s_i;w)}\leq L.\label{eq:smoothness}
\end{align}
\end{assumption}
This assumption is conventional in prior works on policy gradient optimization~\cite{fallah2021convergence,shen2019hessian,papini2018stochastic,agarwal2020optimality,rajeswaran2019meta}. Particularly, one can see that for Softmax policy~\cite{fallah2021convergence}[Appendix~D], which is customary in practice, both \eqref{eq:bounded-grad} and \eqref{eq:smoothness} hold. Moreover, recall that the reward functions $r_i(\cdot,\cdot)$ are nonnegative and bounded, i.e., there exists a constant $R$ such that for all \mbox{$i\in\mcI, a_i\in\mcA_i, r_i\in\mcS_i$}, we have \mbox{$0\leq r_i(s_i,a_i)\leq R$}.

\begin{lemma}[\cite{fallah2021convergence}, Lemma~1, Properties of $J_i$]\label{lem:property-j-i}
Let Assumption~\ref{assump:main} hold. Then, for all $i\in\mcI$ and $w\in\bbR^d$, and any batch of trajectories $\mcD_i$ sampled from distribution $q_i(\cdot;w)$, we have:
\begin{align}
   \big\lVert\nabla J_i(w)\big\rVert, \big\lVert\nabla \hat{J}_i(\mcD_i;w)\big\rVert &\leq \hat{G},\label{eq:bounded-gradient-j-i}\\
   \big\lVert\nabla^2 J_i(w)\big\rVert, \big\lVert\nabla^2 \hat{J}_i(\mcD_i;w)\big\rVert &\leq \hat{L},\label{eq:smoothness-j-i}
\end{align}
where $\hat{G} \coloneqq \tfrac{GR}{(1{-}\gamma)^2}$ and $\hat{L} \coloneqq \tfrac{\left(HG^2+L\right)R}{(1{-}\gamma)^2}$.
\end{lemma}

Lemma~\ref{lem:property-j-i} indicates gradient boundedness and smoothness for the customary value function $J_i$, for each task $i\in\mcI$. Note that this work defines the trajectory up to $H$ actions starting from $s_0$. Hence, the value $\hat{L}$ of is slightly different from~\cite{fallah2021convergence}. By modifying the proof of Lemma~\ref{lem:property-j-i}, we may replace $\frac{1}{1{-}\gamma}$ with $\min\{\frac{1}{1{-}\gamma}, H\}$ due to the fact that the tasks are MDPs with finite time horizons.

\begin{lemma}[Properties of $V_i$]\label{lem:property-v-i}
Let Assumption~\ref{assump:main} hold and $\lambda \geq \kappa \hat{L}$ for some $\kappa{>}1$, and $\hat{G},\hat{L}$ as in Lemma~\ref{lem:property-j-i}. Then, for all $i\in\mcI$ and $w, v\in\bbR^d$, the following properties hold:
\begin{align}
   \big\lVert\nabla V_i(w)\big\rVert &\leq \hat{G},\label{eq:bounded-gradient-v-i}\\
   \big\lVert\nabla V_i(w) - \nabla V_i(v)\big\rVert &\leq \tilde{L}\left\lVert w - v\right\rVert,\label{eq:smoothness-v-i}
\end{align}
where $\tilde{L}\coloneqq\frac{\lambda}{\kappa{-}1}$.
\end{lemma}
This lemma suggests that the Moreau Envelope surrogate value function in \eqref{eq:memrl} has similar properties as the value function in \eqref{eq:cost-j-i}. The upper bound on the gradient is the same, and the smoothness parameter depends on the regularization term $\lambda$. Also note that the global expected value function $V(\cdot)$ has similar boundedness and smoothness properties as each $V_i(\cdot)$, according to the definition in \eqref{eq:memrl-global}.

Now, we state the proof of the above lemma.

\begin{proof}[\textbf{Proof of Lemma~\ref{lem:property-v-i}}]
Before proceeding with the proof, let us present a set of inequalities we will use in the proofs. For a constant \mbox{$\alpha>0$} and set of $m$ vectors \mbox{$\{w_i\}_{i{=}1}^m$} such that \mbox{$w_i\in\bbR^d$}, we have
\allowdisplaybreaks
\begin{subequations}\label{eq:gen-ineq}
\begin{align}
    \lVert w_i + w_j \rVert^2 &\leq (1{+}\alpha)\lVert w_i\rVert^2 + (1{+}\alpha^{-1})\lVert w_j\rVert^2,\label{eq:gen-ineq-1}
    \\\lVert w_i + w_j \rVert &\leq \lVert w_i\rVert + \lVert w_j\rVert,\label{eq:gen-ineq-2}
    \\\left\lVert\sum\limits_{i=1}^m  w_i\right\rVert^2 &\leq m \left(\sum\limits_{i=1}^m \lVert w_i\rVert^2\right)\label{eq:gen-ineq-3},
    \\\left\lVert \bbE\left[w_i\right]\right\rVert &\leq \bbE\left[\left\lVert w_i\right\rVert\right]\label{eq:gen-ineq-4},
    \\-\lVert w_i\rVert^2 {-} \lVert w_j\rVert^2 &\leq 2\left\langle w_i, w_j \right\rangle \leq \lVert w_i\rVert^2 {+} \lVert w_j\rVert^2\label{eq:gen-ineq-5}.
\end{align}
\end{subequations}
To prove the bound on the gradient norm in \eqref{eq:bounded-gradient-v-i}, we have
\begin{align}
\begin{split}
    \left\lVert\nabla V_i(w)\right\rVert \overset{\eqref{eq:grad-v-i}}{=} \left\lVert\lambda(\hat{\theta}_i(w) - w)\right\rVert
    \overset{\eqref{eq:grad-v-i-proof}}{=} \left\lVert\nabla J_i(\hat{\theta}_i(w))\right\rVert \overset{\eqref{eq:bounded-gradient-j-i}}{\leq} \hat{G}.
\end{split}
\end{align}
Moreover, we can show the smoothness in \eqref{eq:smoothness-v-i} by
\begin{align}
    \lVert\nabla &V_i(w)-\nabla V_i(v)\rVert\nonumber\\
    &\overset{\eqref{eq:grad-v-i}}{=} \left\lVert \nabla J_i(\hat{\theta}_i(w)) - \nabla J_i(\hat{\theta}_i(v))\right\rVert\\
    \overset{\eqref{eq:smoothness-j-i}}&{\leq} \hat{L}\left\lVert \hat{\theta}_i(w) - \hat{\theta}_i(v)\right\rVert\\
    \overset{\eqref{eq:grad-v-i-proof}}&{=} \hat{L}\left\lVert \frac{1}{\lambda}\nabla J_i(\hat{\theta}_i(w)) + w - \frac{1}{\lambda}\nabla J_i(\hat{\theta}_i(v)) - v\right\rVert\\
    \overset{\eqref{eq:gen-ineq-2}}&{\leq} \hat{L}\left\lVert w - v\right\rVert + \frac{\hat{L}}{\lambda}\left\lVert \nabla J_i(\hat{\theta}_i(w)) - \nabla J_i(\hat{\theta}_i(v))\right\rVert\\
    \overset{\eqref{eq:grad-v-i}}&{=} \hat{L}\left\lVert w - v\right\rVert + \frac{\hat{L}}{\lambda}\lVert\nabla V_i(w)-\nabla V_i(v)\rVert\\
    \Rightarrow& \lVert\nabla V_i(w)-\nabla V_i(v)\rVert \leq \frac{\lambda \hat{L}}{\lambda {-} \hat{L}} \left\lVert w - v\right\rVert,
\end{align}
where \mbox{$\frac{\lambda \hat{L}}{\lambda {-} \hat{L}} \leq \tilde{L}{\coloneqq}\frac{\lambda}{\kappa{-}1}$}, which completes the proof of \eqref{eq:smoothness-v-i}.
\end{proof}


We are now ready to present our main technical result.

\begin{theorem}[\mtext{MEMRL} Convergence]\label{thm:memrl}
Let Assumption~\ref{assump:main} hold, $\lambda {>} \hat{L}$, and $\alpha=\frac{1}{4\tilde{L}}$. Then for any timestep \mbox{$T\geq 4\tilde{L}^2$}, the following property holds for the iterates of Algorithm~\ref{alg:MEMRL}:
\begin{align*}
     \frac{1}{T}\sum_{t{=}0}^{T{-}1}\lVert\nabla V(w^t)\rVert^2
    \overset{}&{\leq} \frac{8R}{(1{-}\gamma)\sqrt{T}} +\frac{\lambda^2\nu^2}{(\lambda{-}\hat{L})^2} + \frac{8\tilde{L}\hat{G}^2}{B\sqrt{T}}\nonumber\\
    &+ \frac{8\tilde{L}\lambda^2\nu^2}{(\lambda{-}\hat{L})^2B\sqrt{T}} + \frac{8\alpha\tilde{L}\lambda^2\hat{G}^2}{(\lambda{-}\hat{L})^2BD\sqrt{T}},
\end{align*}
where $\hat{G}, \hat{L}$ as in Lemma~\ref{lem:property-j-i}, and $\tilde{L}$ as in Lemma~\ref{lem:property-v-i}.
\end{theorem}

Theorem~\ref{thm:memrl} implies a sublinear convergence rate for \mtext{MEMRL} algorithm to a first-order stationary point with oracle complexity \mbox{$\mcO({1}/{\sqrt{T}}) + \mcO(\nu^2)$}. In other words, to reach a complexity of \mbox{$\mcO(\varepsilon)$}, it is sufficient to run the algorithm for \mbox{$T = \mcO(1/\varepsilon^2)$} iterations via an inexact inner solver with precision $\nu=\mcO(\sqrt{\varepsilon})$.

We are now ready to show the proof of the above Theorem.

\begin{proof}[\textbf{Proof of Theorem~\ref{thm:memrl}}]
Let us start by rewriting the \mtext{MEMRL} method (Alg.~\ref{alg:MEMRL}) by denoting $\tilde{\nabla}V(w^t)$ and $\nabla\tilde{V}(w^t)$ as
\begin{align}
    \tilde{\nabla}V(w^t) &= \frac{1}{B} \sum_{i\in\mcB^t} \nabla \tilde{V}_i(w^t),\label{eq:stoch-task-batch-grad-v}\\
    \nabla\tilde{V}(w^t) &= \bbE_{i\sim p}\left[\nabla \tilde{V}_i(w^t)\right]\label{eq:stoch-grad-v},
\end{align}
where~\eqref{eq:stoch-grad-v} is the expectation of~\eqref{eq:stoch-task-batch-grad-v}.
Therefore, we obtain the following update rule
\begin{align}\label{eq:update-rule}
    w^{t{+}1} = w^t + \alpha \tilde{\nabla} V(w^t),
\end{align}
for \mtext{MEMRL}. Then, according to Lemma~\ref{lem:property-v-i}, we have
\begin{align}
    \big| V(w^{t{+}1}) &- V(w^t) - \big\langle \nabla V(w^t), w^{t{+}1} - w^t\big\rangle\big|\\
    &\qquad\qquad\qquad\qquad\qquad\leq \frac{\tilde{L}}{2}\left\lVert w^{t{+}1} - w^t \right\rVert^2\overset{\eqref{eq:update-rule}}{\Rightarrow}\nonumber\\
    -V(w^{t{+}1}) &\leq -V(w^t) - \alpha\underbrace{\big\langle \nabla V(w^t), \tilde{\nabla}V(w^t)\big\rangle}_{A_1}\nonumber\\
    &\qquad\qquad\quad + \frac{\alpha^2\tilde{L}}{2} \underbrace{\big\lVert\tilde{\nabla}V(w^t)\big\rVert^2}_{A_2},\label{eq:main-ineq-L}
\end{align}

where by taking conditional expectation from \eqref{eq:main-ineq-L} conditioned on $\mcF^{t}$, we have
\begin{align}
    -\bbE\left[V(w^{t{+}1})|\mcF^t\right] \leq -V(w^t) - \bbE\left[\alpha A_1 - \frac{\alpha^2\tilde{L}}{2} A_2\Bigg|\mcF^t\right].\label{eq:main-ineq-L-cond}
\end{align}
Therefore, it is sufficient to show appropriate bounds for the conditional expectation of $A_1$ and $A_2$. First, we have
\begin{align}
    \bbE\left[A_1|\mcF^t\right] &= \bbE\left[\left\langle\nabla V(w^t), \tilde{\nabla}V(w^t)\right\rangle\big|\mcF^t\right]\label{eq:A1-cond-start}\\
     \overset{\eqref{eq:stoch-grad-v}}&{=} \bbE\left[\left\langle\nabla V(w^t), \nabla\tilde{V}(w^t)\right\rangle\big|\mcF^t\right]\\
    &= \lVert\nabla V(w^t)\rVert^2\\
    &+ \bbE\left[\left\langle\nabla V(w^t), \nabla\tilde{V}(w^t) - \nabla V(w^t)\right\rangle\big|\mcF^t\right]\nonumber\\
    \overset{\eqref{eq:gen-ineq-5}}&{\geq} \lVert\nabla V(w^t)\rVert^2 - \frac{1}{2}\lVert\nabla V(w^t)\rVert^2\\
    &-\frac{1}{2} \left\lVert\bbE\left[\nabla\tilde{V}(w^t) - \nabla V(w^t)\big|\mcF^t\right]\right\rVert^2\nonumber\\
    &= \frac{1}{2}\lVert\nabla V(w^t)\rVert^2\\
    &-\frac{1}{2} \left\lVert\bbE_{i\sim p}\left[\bbE\left[\nabla\tilde{V}_i(w^t) - \nabla V_i(w^t)\big|\mcF^t\right]\right]\right\rVert^2.\nonumber
    \\ \overset{\eqref{eq:gen-ineq-4}}&{\geq} \frac{1}{2}\lVert\nabla V(w^t)\rVert^2\label{eq:A1-cond-end}\\
    &-\frac{1}{2} \Bigg(\bbE_{i\sim p}\Big[\Big\lVert\underbrace{\bbE\left[\nabla\tilde{V}_i(w^t) - \nabla V_i(w^t)\big|\mcF^t\right]}_{A_3}\Big\rVert\Big]\Bigg)^2.\nonumber
\end{align}
Now, let us define $F_i(\theta_i;w)$ as
\begin{align}
    F_i(\theta_i;w) &\coloneqq J_i(\theta_i) - \frac{\lambda}{2}\left\lVert\theta_i-w\right\rVert^2 \Rightarrow\\
    \nabla F_i(\theta_i;w) &\coloneqq \nabla J_i(\theta_i) - \lambda(\theta_i-w),\label{eq:grad-f-i}
\end{align}
then due to \eqref{eq:stoch-grad-f-i}, \eqref{eq:grad-f-i}, and the fact that \mbox{$\nabla F_i(\hat{\theta}_i(w);w)=0$},
\begin{align}
    \lVert A_3 \rVert &= \lambda\left\lVert\bbE\left[\tilde{\theta}_i(w^t)-\hat{\theta}_i(w^t) \big|\mcF^t\right]\right\rVert\label{eq:def-A3}\\
    &=\Big\lVert\bbE\Big[\nabla \tilde{J}_i\left(\mcD_i^t;\tilde{\theta}_i(w^t)\right)-\nabla\tilde{F}_i\left(\mcD_i^t;\tilde{\theta}_i(w^t),w^t\right)\nonumber\\
    &\qquad\qquad\qquad\qquad\qquad\,\,-\nabla J_i(\hat{\theta}_i(w^t)) \big|\mcF^t\Big]\Big\rVert\\
    \overset{\eqref{eq:gen-ineq-2}}&{\leq}\Big\lVert\bbE\Big[\nabla J_i(\tilde{\theta}_i(w^t))-\nabla J_i(\hat{\theta}_i(w^t)) \big|\mcF^t\Big]\Big\rVert + \nu\\
    \overset{\eqref{eq:smoothness-j-i}}&{\leq}\hat{L}\Big\lVert\bbE\Big[\tilde{\theta}_i(w^t)-\hat{\theta}_i(w^t)\big|\mcF^t\Big]\Big\rVert + \nu
    \\
    \overset{\eqref{eq:def-A3}}&{=} \frac{\hat{L}}{\lambda}\lVert A_3 \rVert+ \nu \Rightarrow\\
    \lVert A_3 \rVert &\leq \frac{\lambda}{\lambda{-}\hat{L}}\nu.\label{eq:def-A3-end}
\end{align}
Therefore, according to \eqref{eq:A1-cond-start}-\eqref{eq:A1-cond-end} and \eqref{eq:def-A3}-\eqref{eq:def-A3-end}, we have
\begin{align}
    \bbE\left[A_1|\mcF^t\right] \geq \frac{1}{2}\lVert\nabla V(w^t)\rVert^2 - \frac{\lambda^2\nu^2}{2(\lambda{-}\hat{L})^2}.\label{eq:A1-final}
\end{align}
We now bound the expectation of $A_2$ conditioned on $\mcF^t$.
\begin{align}
    \bbE\left[A_2|\mcF^t\right] \overset{\eqref{eq:gen-ineq-3}}&{\leq} 2\lVert\nabla V(w^t)\rVert^2\label{eq:A2-cond-start}\\
    &+ 2\bbE\left[\left\lVert \tilde{\nabla}V(w^t) - \nabla V(w^t)\right\rVert^2\big|\mcF^t\right]\nonumber\\
    &= 2\lVert\nabla V(w^t)\rVert^2\label{eq:A2-cond-end}\\
    &+ 2\bbE\Big[\underbrace{\bbE_{i\sim p}\big\lVert\tilde{\nabla}V(w^t) - \nabla V(w^t)\big\rVert^2}_{A_4}\big|\mcF^t\Big].\nonumber
\end{align}
We can bound $A_4$ using Lemma~\ref{lem:property-v-i} and conditional independence. So,
\begin{align}
    \bbE&[A_4|\mcF^t] = \bbE\left[\bbE_{i}\left\lVert\frac{1}{B}\sum_{i\in\mcB^t}\nabla\tilde{V}_i(w^t) - \nabla V(w^t)\right\rVert^2\Big|\mcF^t\right]\\
    \overset{\eqref{eq:gen-ineq-3}}&{\leq} \frac{2}{B^2}\bbE\left[\bbE_{i}\left\lVert\sum_{i\in\mcB^t}\left[\nabla\tilde{V}_i(w^t)-\nabla V_i(w^t)\right]\right\rVert^2\Big|\mcF^t\right]\nonumber\\
    &+\frac{2}{B^2}\bbE\left[\bbE_{i}\left\lVert\sum_{i\in\mcB^t}\left[\nabla V_i(w^t) - \nabla V(w^t)\right]\right\rVert^2\Big|\mcF^t\right]\\
    \overset{}&{\leq} \frac{2}{B}\bbE\left[\bbE_{i}\left\lVert\nabla\tilde{V}_i(w^t)-\nabla V_i(w^t)\right\rVert^2\big|\mcF^t\right]\nonumber\\
    &+\frac{2}{B}\bbE\left[\bbE_{i}\left\lVert\nabla V_i(w^t) - \nabla V(w^t)\right\rVert^2\big|\mcF^t\right]\label{eq:A4-full-grad-independent-batch}\\
    \overset{\eqref{eq:bounded-gradient-v-i}}&{\leq} \frac{2}{B}\bbE\Big[\bbE_{i}\underbrace{\big\lVert\nabla\tilde{V}_i(w^t)-\nabla V_i(w^t)\big\rVert^2}_{A_5}\big|\mcF^t\Big] + \frac{8\hat{G}^2}{B},\label{eq:A4-full-grad-hat-G}
\end{align}
where \eqref{eq:A4-full-grad-independent-batch} holds due to the conditional independency between the summation terms and \eqref{eq:A4-full-grad-hat-G} holds according to \eqref{eq:bounded-gradient-v-i} in Lemma~\ref{lem:property-v-i}. It is sufficient to show a bound for $A_5$ using techniques such as \eqref{eq:def-A3}-\eqref{eq:def-A3-end}. Before showing the bound, note that $F_i(\theta_i;w)$ is a $(\lambda{-}\hat{L})$-smooth function with respect to the auxiliary parameter $\theta_i$ according to~\cite{toghani2022persafl,dinh2020personalized}. Hence,
\begin{align}
    A_5 &= \lambda^2\left\lVert\tilde{\theta}_i(w^t)-\hat{\theta}_i(w^t)\right\rVert^2\label{eq:def-A5}\\
    \overset{\text{\tiny smooth}}&{\leq} \frac{\lambda^2}{(\lambda{-}\hat{L})^2} \Big\lVert\nabla F_i\big(\tilde{\theta}_i(w^t);w^t\big)-\underbrace{\nabla F_i\big(\hat{\theta}_i(w^t);w^t\big)}_{=0}\Big\rVert^2\\
    &= \frac{\lambda^2}{(\lambda{-}\hat{L})^2} \Big\lVert\nabla F_i\big(\tilde{\theta}_i(w^t);w^t\big) - \nabla \tilde{F}_i\big(\mcD_i^t;\tilde{\theta}_i(w^t),w^t\big)\nonumber\\
    &\qquad\qquad+ \nabla \tilde{F}_i\big(\mcD_i^t;\tilde{\theta}_i(w^t),w^t\big)\Big\rVert^2\\
    \overset{\eqref{eq:gen-ineq-3}}&{\leq} \frac{2\lambda^2}{(\lambda{-}\hat{L})^2}\Bigg[\Big\lVert\nabla F_i\big(\tilde{\theta}_i(w^t);w^t\big) - \nabla \tilde{F}_i\big(\mcD_i^t;\tilde{\theta}_i(w^t),w^t\big)\Big\rVert^2\nonumber\\
    &+ \Big\lVert\nabla\tilde{F}_i\big(\mcD_i^t;\tilde{\theta}_i(w^t),w^t\big)\Big\rVert^2\Bigg]\\
    &= \frac{2\lambda^2}{(\lambda{-}\hat{L})^2}\Bigg[\underbrace{\Big\lVert\nabla J_i\big(\tilde{\theta}_i(w^t)\big) - \nabla \tilde{J}_i\big(\mcD_i^t;\tilde{\theta}_i(w^t)\big)\Big\rVert^2}_{A_6} + \nu^2\Bigg],\label{eq:def-A5-end}
\end{align}
where due to the conditional independency between the trajectories, we have
\begin{align}
    \bbE[A_6|\mcF^t] \leq \frac{\hat{G}^2}{D}.\label{eq:A6}
\end{align}
Therefore, from \eqref{eq:A2-cond-start}-\eqref{eq:A6}, we can conclude that
\begin{align}
    \bbE\left[A_2|\mcF^t\right] &\leq 2\bbE\lVert\nabla V(w^t)\rVert^2 + \frac{8\lambda^2\hat{G}^2}{(\lambda{-}\hat{L})^2BD}\nonumber\\
    &+ \frac{8\lambda^2\nu^2}{(\lambda{-}\hat{L})^2B} + \frac{8\hat{G}^2}{B}.\label{eq:A2-final}
\end{align}
So finally, according to \eqref{eq:main-ineq-L-cond}, \eqref{eq:A1-final}, and \eqref{eq:A2-final}, we have
\begin{align}
    -\bbE\left[V(w^{t{+}1})|\mcF^t\right] &\leq -V(w^t) - \frac{\alpha}{2}(1{-}2\alpha \tilde{L})\,\bbE\lVert\nabla V(w^t)\rVert^2\nonumber\\
    &+\frac{\alpha\lambda^2\nu^2}{2(\lambda{-}\hat{L})^2} + \frac{4\alpha^2\tilde{L}\lambda^2\hat{G}^2}{(\lambda{-}\hat{L})^2BD}\nonumber\\
    &+ \frac{4\alpha^2\tilde{L}\lambda^2\nu^2}{(\lambda{-}\hat{L})^2B} + \frac{4\alpha^2\tilde{L}\hat{G}^2}{B}
    .\label{eq:ineq-alpha}
\end{align}
Then, by taking an expectation from \eqref{eq:main-ineq-alpha} and averaging the inequality for $t=0,1,\dots,T{-}1$, under the assumption of $\alpha\leq\frac{1}{4\tilde{L}}$, we have
\begin{align}
     \frac{1}{T}\sum_{t{=}0}^{T{-}1}\bbE\lVert\nabla V(w^t)&\rVert^2\leq \frac{4}{\alpha T}\bbE\left[V(w^{T})-V(w^0)\right]\nonumber\\
    &+\frac{\lambda^2\nu^2}{(\lambda{-}\hat{L})^2} + \frac{16\alpha\tilde{L}\lambda^2\hat{G}^2}{(\lambda{-}\hat{L})^2BD}\nonumber\\
    &+ \frac{16\alpha\tilde{L}\lambda^2\nu^2}{(\lambda{-}\hat{L})^2B} + \frac{16\alpha\tilde{L}\hat{G}^2}{B}\\
    \overset{}&{\leq} \frac{4R}{\alpha T(1{-}\gamma)} +\frac{\lambda^2\nu^2}{(\lambda{-}\hat{L})^2} + \frac{16\alpha\tilde{L}\lambda^2\hat{G}^2}{(\lambda{-}\hat{L})^2BD}\nonumber\\
    &+ \frac{16\alpha\tilde{L}\lambda^2\nu^2}{(\lambda{-}\hat{L})^2B} + \frac{16\alpha\tilde{L}\hat{G}^2}{B},\label{eq:main-ineq-alpha}
\end{align}

where \eqref{eq:main-ineq-alpha} holds since the reward value is bounded between $0$ and $R$. By setting \mbox{$\alpha=\frac{1}{2\sqrt{T}}$}, for \mbox{$T\geq 4\tilde{L}^2$}, we can guarantee \mbox{$\alpha\leq\frac{1}{4\tilde{L}}$} as well as
\begin{align}
     \frac{1}{T}\sum_{t{=}0}^{T{-}1}\lVert\nabla V(w^t)\rVert^2
    \overset{}&{\leq} \frac{8R}{(1{-}\gamma)\sqrt{T}} +\frac{\lambda^2\nu^2}{(\lambda{-}\hat{L})^2} + \frac{8\tilde{L}\hat{G}^2}{B\sqrt{T}}\nonumber\\
    &+ \frac{8\tilde{L}\lambda^2\nu^2}{(\lambda{-}\hat{L})^2B\sqrt{T}} + \frac{8\alpha\tilde{L}\lambda^2\hat{G}^2}{(\lambda{-}\hat{L})^2BD\sqrt{T}},\label{eq:main-ineq-T}
\end{align}
which completes the proof of Theorem~\ref{thm:memrl}.
\end{proof}


\begin{figure*}
    {\centering
    \includegraphics[width=0.28\linewidth]{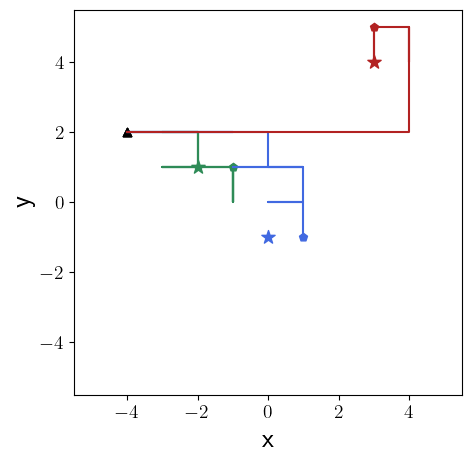}
    \includegraphics[width=0.28\linewidth]{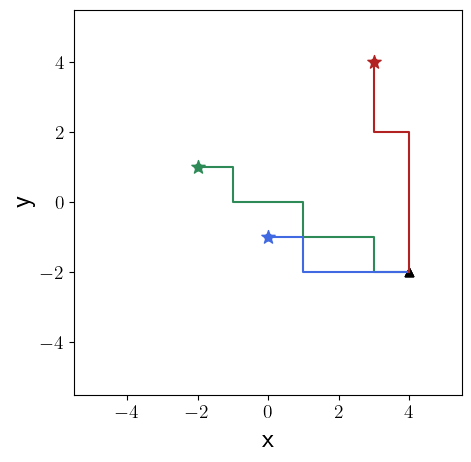}
    \includegraphics[width=0.42\linewidth]{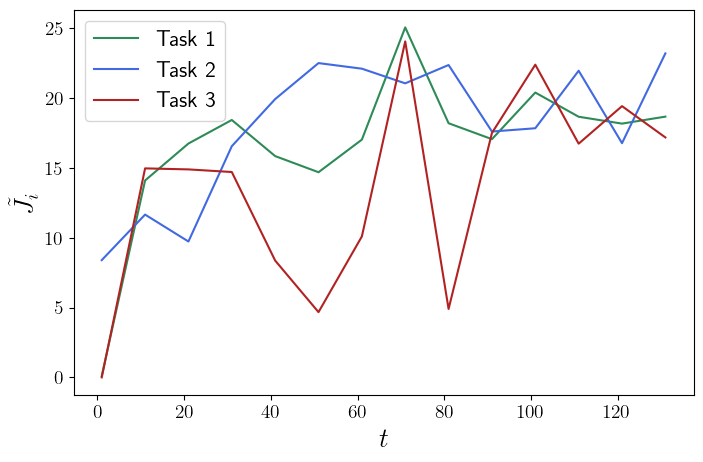}}
    
    \caption{The performance of our \mtext{MEMRL} algorithm on discrete $2$D-navigation for $|\mcI|{=}3$ tasks with different underlying MDPs.} \textbf{(Left)} The navigation map at iteration $t=0$ starting from a random location (black triangle) on the grid. The stars indicate the destination of each task $i\in\mcI$. Pentagons indicate the end of a trajectory when it fails to reach its destination (star). \textbf{(Middle)} The navigation map at iteration $t=120$, where the adapted meta-policy for each task is optimal. \textbf{(Right)} The evolution of individual reward functions given the adapted meta-policy on each task. Each curve is the empirical mean of the reward obtain over $10$ independent trajectories conditioned on the approximated policy parameter $\tilde{\theta}_i^t$.
    \label{fig:memrl-2d-navigation}
\end{figure*}

\section{Numerical Experiment}\label{sec:experiments}
In this section, we present numerical studies of our proposed \mtext{MEMRL} algorithm. We consider a discrete variation of $2$D-navigation problem \cite{henderson2017multitask,fallah2021convergence,rothfuss2018promp,finn2017model} over a square grid, where the objective is to reach a specified destination by taking valid actions. The \mtext{MRL} setup for this problem narrates the scenario where the goal is to obtain a meta policy that can be easily adapted to perform well for multiple destinations.

Now, let us describe the problem setup. We consider a group of $|\mcI|=3$ tasks where the objective of each task $i\in\mcI$ is to navigate to some corresponding destination $s_i^\star = \{x_i, y_i\}$ over an $11 \times 11$ grid, $\{-5, \dots, 5\} \times \{-5, \dots, 5\}$, starting from some random point. For example, check the left subfigure in Figure~\ref{fig:memrl-2d-navigation}, where each star represents the destination locations for some task $i\in[3]$. The set of valid actions is limited to left, right, down, up, and pause,
\begin{align}
    \mcA_i = \{(0, 0), (0, 1), (0, -1), (1, 0), (-1, 0)\}.
\end{align}
The set of states \mbox{$\mS_i=\{-5, \dots, 5\} \times \{-5, \dots, 5\}$}. We define the reward function inversely proportional~\cite{rothfuss2018promp} to the $\ell_1$ distance of the agent's next state to the intended destination, i.e., 
\begin{align}
    r_i(a_i^h|s_i^h)=\exp(-{\lVert s_i^{h{+}1}-s_i^\star\rVert}_1),
\end{align}
where $s_i^{h{+}1}$ is the next coordinate of the agent after taking action $a_i^h$. Finally, we parameterize the policy with a two-layer \mtext{MLP} network with $\mathrm{softmax}$ layer and $5$ outputs by taking a two-dimensional input, i.e., the state of the agent yields a probability vector of the potential actions.

We implement Step~\ref{step:approx} of Algorithm~\ref{alg:MEMRL} via the first-order inexact optimizer which we described in Section~\ref{sec:setup} under fixed inner loop with $K{=}8$ steps. Moreover, we select $\lambda=2$, $\alpha=0.1$, $\beta=0.02$, $\gamma=0.99$, $B=2$, and $D=10$. The right subfigure of Figure~\ref{fig:memrl-2d-navigation} illustrates the performance of \mtext{MEMRL} for the $2$D-navigation problem. We plot the stochastic reward function $\tilde{J}_i$ for each task across the optimization runtime. Moreover, we show present the underlying navigation before and after training with out method respectively in the left and middle subfigures of Figure~\ref{fig:memrl-2d-navigation}.


\section{Conclusions}
\label{sec:conclusion}

We studied the Meta-Reinforcement Learning (\mtext{MRL}) problem and introduced \mtext{MEMRL}. This novel meta-reinforcement learning algorithm leverages Moreau Envelopes to achieve fast and stable policy adaptation with first-order optimization. We proved the convergence of our algorithm under non-convex policy gradient optimization and showed its performance over a discrete $2$D-navigation problem with no need to second-order information. Our work opens up new directions for applying Moreau Envelopes to other meta-learning problems to resolve the scalability challenges for Hessian computation. A thorough study of the Multi-Agent \mtext{MRL} problem remains a future study for this work. We also leave the extended analysis for infinite horizon MDPs to future studies.


\bibliographystyle{IEEEtran}
\bibliography{ref}

\end{document}